\documentclass[runningheads]{llncs}

\usepackage{graphicx}
\usepackage{amsmath}
\usepackage{hyperref}
\hypersetup{
    colorlinks=true,    
    urlcolor=blue,
}
\usepackage{multirow}
\usepackage[table]{xcolor}
\definecolor{mygreen}{rgb}{0.631,0.800,0.553}
\definecolor{myyellow}{rgb}{1,.85,.55}
\definecolor{myred}{rgb}{.97, .68, .55}
\definecolor{myblue}{rgb}{.61, .75, .90}
\usepackage{mdframed}
\usepackage{boldline}
\usepackage{dsfont}

\begin{document}

\title{Feature Extraction for Generative Medical Imaging Evaluation: New Evidence Against an Evolving Trend}
\titlerunning{Feature Extraction for Generative Medical Imaging Evaluation}
\author{McKell Woodland\inst{1,2} \and
        Austin Castelo\inst{1} \and
        Mais Al Taie\inst{1} \and
        Jessica Albuquerque Marques Silva\inst{1} \and
        Mohamed Eltaher\inst{1} \and
        Frank Mohn\inst{1} \and
        Alexander Shieh\inst{1} \and
        Suprateek Kundu\inst{1} \and
        Joshua P. Yung\inst{1} \and
        Ankit B. Patel\inst{2,3} \and
        Kristy K. Brock\inst{1}}
\authorrunning{M. Woodland et al.}
\institute{The University of Texas MD Anderson Cancer Center, Houston TX 77030, USA            \email{MEWoodland@mdanderson.org} \and
           Rice University, Houston TX 77005, USA \and
           Baylor College of Medicine, Houston TX 77030, USA}
\maketitle

\begin{abstract}
Fr\'{e}chet Inception Distance (FID) is a widely used metric for assessing synthetic image quality.
It relies on an ImageNet-based feature extractor, making its applicability to medical imaging unclear.
A recent trend is to adapt FID to medical imaging through feature extractors trained on medical images.
Our study challenges this practice by demonstrating that ImageNet-based extractors are more consistent and aligned with human judgment than their RadImageNet counterparts.
We evaluated sixteen StyleGAN2 networks across four medical imaging modalities and four data augmentation techniques with Fr\'{e}chet distances (FDs) computed using eleven ImageNet or RadImageNet-trained feature extractors.
Comparison with human judgment via visual Turing tests revealed that ImageNet-based extractors produced rankings consistent with human judgment, with the FD derived from the ImageNet-trained SwAV extractor significantly correlating with expert evaluations.
In contrast, RadImageNet-based rankings were volatile and inconsistent with human judgment.
Our findings challenge prevailing assumptions, providing novel evidence that medical image-trained feature extractors do not inherently improve FDs and can even compromise their reliability.
Our code is available at \url{https://github.com/mckellwoodland/fid-med-eval}.

\end{abstract}

\section{Introduction}

Fréchet Inception Distance (FID) is the most commonly used metric for evaluating synthetic image quality \cite{Heusel2017}. 
It quantifies the Fr\'echet distance (FD) between two Gaussian distribution curves fitted to embeddings of real and generated images.
These embeddings are typically extracted from the penultimate layer of an InceptionV3 network trained on ImageNet.
FID's utility has been demonstrated through its correlation with human judgment \cite{Woodland2022}, sensitivity to distortions \cite{Heusel2017}, capability to detect overfitting \cite{Borji2019}, and relative sample efficiency \cite{Borji2019}.
Nonetheless, the metric has faced criticism, including that the InceptionV3 network may only embed information relevant to ImageNet class discrimination \cite{Truong2021,kynkäänniemi2023}.

Three approaches exist for adapting FID to medical imaging.
The first involves using an InceptionV3 extractor trained on a large, publicly available medical dataset, such as RadImageNet, a database containing 1.35 million annotated computed tomography (CT), magnetic resonance imaging (MRI), and ultrasonography exams \cite{Mei2022,Osuala2023_medigan}.
While a RadImageNet-based FD considers medically relevant features, its efficacy remains largely unexplored.
One potential bias is that networks trained for disease detection may focus too heavily on small, localized regions \cite{Anton2022} to effectively evaluate an entire image's quality. 
Additionally, RadImageNet-based FDs may not generalize to new medical modalities \cite{Osuala2023_medigan} or patient populations.
Our novel comparison of RadImageNet-base FDs to human judgment revealed discrepancies, even on in-domain abdominal CT data.

The second approach utilizes self-supervised networks for feature extraction \cite{Morozov2020}.
These networks are encouraging as they create transferable and robust representations \cite{He2020}, including on medical images \cite{Truong2021}.
Despite their promise, the lack of publicly available, self-supervised models trained on extensive medical imaging datasets has hindered their application.
Our study is the first to employ self-supervised extractors for synthetic medical image evaluation.
We find a significant correlation between an FD derived from an ImageNet-trained SwAV network (FSD) and medical experts' appraisal of image realism, highlighting the potential of self-supervision for advancing generative medical imaging evaluation.

The third approach employs a feature extractor trained on the dataset used to train the generative imaging model \cite{Chen2021,Jung2021,Tronchin2021}. 
While advantageous for domain coherence, the algorithm designer essentially creates the metric used to evaluate their algorithm, resulting in unquantified bias \cite{Woodland2022}. 
Moreover, the private and varied nature of these feature extractors poses challenges for reproducibility and benchmarking. 
Given these limitations, our study focused on publicly available feature extractors.

Our study offers a novel comparison of generative model rankings created by ImageNet and RadImageNet-trained feature extractors with expert judgment.
Our key contributions are: 
\begin{enumerate}
    \item Demonstrating that ImageNet-based feature extractors consistently produce more realistic model rankings than their RadImageNet-based counterparts.
    This finding raises concerns about the prevalent practice of using medical image-trained feature extractors for generative model ranking without evaluating the efficacy of the proposed metric.
    \item Identifying a significant correlation between an FD calculated with an ImageNet-trained SwAV network and expert assessments of image realism, demonstrating that FSD is a viable alternative to FID on medical images.
    \item Benchmarking multiple data augmentation techniques designed to enhance generative performance within limited data domains on medical imaging datasets.
    \item Introducing a novel method for evaluating visual Turing Tests (VTTs) via hypothesis testing, providing an unbiased measure of participant perception of synthetic image realism.
\end{enumerate}

\section{Methods}

\subsection{Generative Modeling}

Four medical imaging datasets were used for generative modeling: the Segmentation of the Liver Competition 2007 (SLIVER07) dataset with 20 liver CT studies \cite{Heimann2009}\footnote{\url{https://sliver07.grand-challenge.org/}}, the ChestX-ray14 dataset with 112,100 chest X-rays \cite{Wang2017}\footnote{\url{https://nihcc.app.box.com/v/ChestXray-NIHCC}}, the brain tumor dataset from the Medical Segmentation Decathlon (MSD) with 750 brain MRI studies \cite{Antonelli2022,Simpson2019}\footnote{\url{http://medicaldecathlon.com/}, CC-BY-SA 4.0 license.}, and the Automated Cardiac Diagnosis Challenge (ACDC) dataset with 150 cardiac cine-MRIs \cite{Bernard2018}\footnote{\url{https://www.creatis.insa-lyon.fr/Challenge/acdc/databases.html}}.
Multi-dimensional images were converted to two dimensions by extracting axial slices and excluding the slices with less than 15\% nonzero pixels.

To enable a comparison of synthetic quality, four StyleGAN2 \cite{Karras2020_CVPR} models were trained per dataset, using either adaptive discriminator augmentation (ADA) \cite{Karras2020_NeurIPS}, differentiable augmentation (DiffAugment) \cite{Zhao2020}, adaptive pseudo augmentation (APA) \cite{Jiang2021}, or no augmentation.
While all of the data augmentation techniques were created to improve the performance of generative models on limited data domains, such as medical imaging, we are the first to benchmark the techniques on medical images.
Each model was evaluated using the weights obtained at the end of 25,000 kimg (a kimg represents a thousand real images being shown to the discriminator), except for the MSD experiments, which were limited to 5,000 kimg due to training instability.
Our code and trained model weights are available at \url{https://github.com/mckellwoodland/fid-med-eval}. 

\subsection{Human Evaluation}

Human perception of model quality was assessed with one VTT per model.
Each test comprised 20 randomly selected images with an equal number of real and generated images.
Participants were asked to identify whether each image was real or generated and rate its realism on a Likert scale from 1 to 3 (1: “Not at all realistic,” 2: “Somewhat realistic,” and 3: “Very realistic”). 
The tests were administered to five specialists with medical degrees.
In addition to the VTTs, three radiologists were shown 35 synthetic radiographs per ChestX-ray14 model and were asked to rank and provide a qualitative assessment of the models.

False positive rate (FPR) and false negative rate (FNR) were used to evaluate the VTTs.
The FPRs represent the proportion of generated images that participants considered to be real.
FPRs near 50\% indicate random guessing.
One-sided paired \textit{t} tests were performed on the FPRs with $\alpha$=.05 to benchmark the data augmentation techniques.
For each VTT, the average Likert ratings of real and generated images were computed per participant.
The difference between these average ratings was then computed to compare the perceived realism of real and generated images. 
Two-sample Kolmogorov-Smirnov (KS) tests were conducted on the Likert ratings of the real and generated images with $\alpha$=.10 to determine whether the ratings came from the same distribution, indicating that the participants viewed the realism of the generated images to be equivalent to that of the real images.
We are the first to use the difference in average Likert ratings and the KS test for generative modeling evaluation.

When taking a VTT, participants may be more likely to select either ``real'' or ``generated'' when uncertain.
This bias causes the average FPR to not fully encapsulate whether participants could differentiate between real and generated images.
To address this challenge, we propose a novel method for evaluating VTTs via hypothesis testing. 
The method aims to demonstrate that the likelihood of a participant selecting ``real'' is the same for both real and generated images. 
For each participant $p$, we define the null hypothesis $\mathds{P}(p\text{ guesses real } | \text{ G})=\mathds{P}(p\text{ guesses real } | \text{ R})$ where G represents the event that the image is generated and R represents the event that the image is real.
We evaluate this hypothesis using a two-sample \textit{t} test with $\alpha$=.10, where the first sample is the participant’s binary predictions for generated images, and the second is their predictions for real images. 
To evaluate VTTs for multiple participants $P$, we define the null hypothesis $\mathds{P}(\text{random } p\in P \text{ guesses real }|\text{ G})=\mathds{P}(\text{random } p\in P\text{ guesses real }|\text{ R})$.
We evaluate this hypothesis via a two-sample \textit{t} test with $\alpha$=.10, where the first sample is the FPR and the second is the true positive rate of each participant. 

\subsection{Fr\'{e}chet Distances}

Quantitative evaluation of synthetic image quality was performed by calculating the FD ${d(\Sigma_1, \Sigma_2, \mu_1, \mu_2)}^2={|\mu_1-\mu_2|}^{2}+\text{tr}(\Sigma_1+\Sigma_2-{2(\Sigma_1\Sigma_2)}^\frac{1}{2})$ \cite{Dowson1982} between two multivariate Gaussians ($\Sigma_R,\mu_R$) and ($\Sigma_G,\mu_G$) fitted to real and generated features extracted from the penultimate layer of eleven backbone networks: InceptionV3 \cite{Szegedy2015}, ResNet50 \cite{He2016}, InceptionResNetV2 \cite{Szegedy2017}, and DenseNet121 \cite{Huang2017} each trained separately on both ImageNet \cite{Deng2009} and RadImageNet \cite{Mei2022}, along with SwAV \cite{Caron2020}, DINO \cite{Caron2021}, and a Swin Transformer \cite{Liu2021} trained on ImageNet.
The first four networks were included to compare all publicly available RadImageNet models to their ImageNet equivalents.
SwAV and DINO were included to evaluate the impact of self-supervision, as self-supervised representations have demonstrated superior transferability to new domains \cite{He2020} and richer embeddings on medical images \cite{Truong2021}. 
Finally, a Swin Transformer \cite{Liu2021} was included as transformers have been shown to create transferable and robust representations \cite{Zhou2021}.
We are the first to use self-supervised and transformer architectures with FD for generative medical imaging evaluation.
As the scale of FDs varies substantially by feature extractor, relative FDs (rFDs) 
$\frac{{d(\Sigma_R, \Sigma_G, \mu_R, \mu_G)}^2}{{d(\Sigma_{R_1},\Sigma_{R_2},\mu_{R_1},\mu_{R_2})}^2}$
were computed with a random split of the real features into two Gaussian distributions ($\Sigma_{R_1},\mu_{R_1}$) and ($\Sigma_{R_2},\mu_{R_2}$).
Paired \textit{t} tests with $\alpha$=0.05 were conducted on the FDs to benchmark the data augmentation techniques.  
The Pearson correlation coefficient with $\alpha$=0.05 was used to quantify the correspondence between the FDs and human judgment and the correspondence between individual FDs.
We are the first to consider whether medical image-based FDs are correlated with human judgment.

\section{Results}

\begin{figure}
    \includegraphics[width=\textwidth]{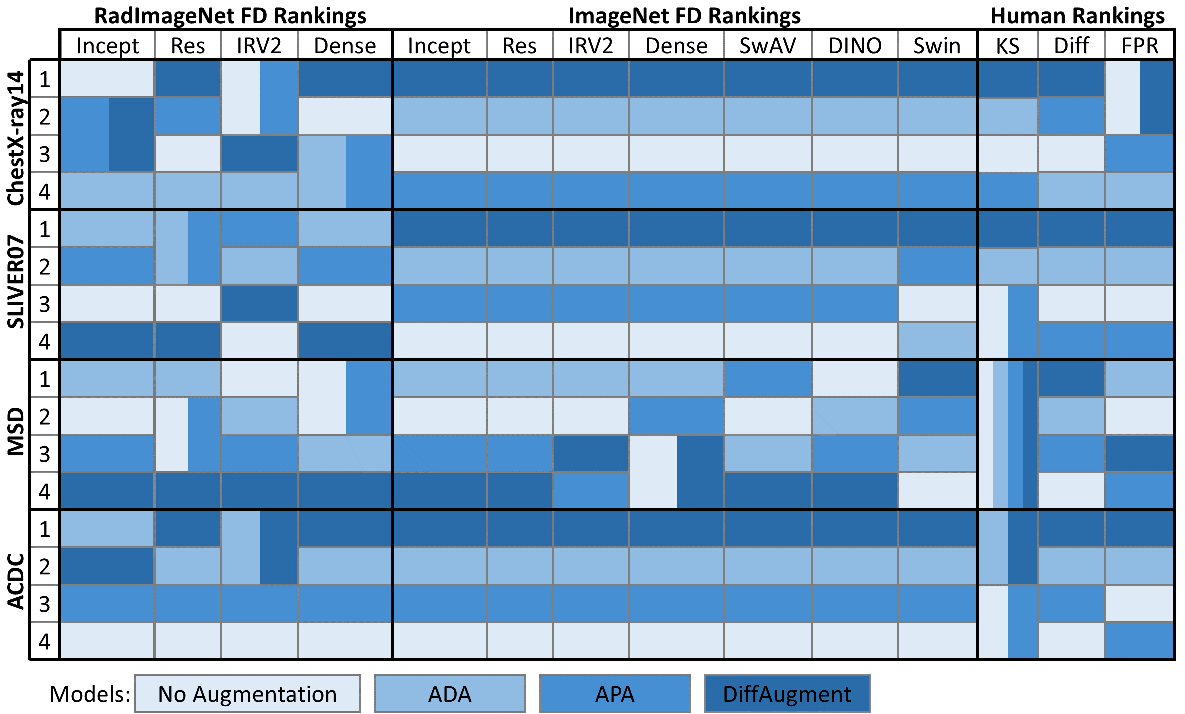}
    \caption{
    Model rankings listed in descending order of performance. 
    FDs are split by dataset and architecture: InceptionV3 (Incept), ResNet50 (Res), InceptionResNetV2 (IRV2), DenseNet121 (Dense), SwAV, DINO, and Swin Transformer (Swin).
    Human rankings are KS test p-values (KS), average difference in mean Likert scores (Diff), and FPRs.
    }
    \label{fig:rank}
\end{figure}

Table \ref{tab:vtt} summarizes the overall results of the VTTs, with detailed individual participant outcomes available at \url{https://github.com/mckellwoodland/fid-med-eval}. 
The rFDs based on ImageNet and RadImageNet are outlined in Tables \ref{tab:rfd_in} and \ref{tab:rfd_rin}, while the FDs can be found in Tables \ref{tab:fd_in} and \ref{tab:fd_rin} in the Appendix. 
Model rankings based on individual metrics are illustrated in Figure \ref{fig:rank}. 
Our analysis revealed

\begin{table}
    \caption{VTT results. 
    Column 1 lists each tested dataset, while Column 2 specifies the augmentation technique (Aug) utilized during model training: no augmentation (None), ADA, APA, and DiffAugment (DiffAug). 
    Columns 3 and 4 showcase the average FPRs and FNRs. 
    FPRs near 50\% imply random guessing. 
    Column 5 provides \textit{t} test p-values, whose null hypothesis is that the probability of a random participant selecting ``real'' is the same for real and generated images. 
    Column 6 displays the average difference between mean Likert ratings for real and generated images (Diff); a negative value indicates that the generated images were perceived to be more realistic than the actual images. 
    Column 7 presents KS test p-values, whose null hypothesis is that the Likert ratings for real and generated images were drawn from the same distribution. 
    $\uparrow$ and $\downarrow$ denote preferable higher or lower values. 
    The underlined boldface type represents the best performance per dataset. 
    Gray boxes indicate failure to reject the null hypothesis, suggesting that participants viewed real and generated images to be equivalent. 
    \textdagger\phantom{ }indicates decreased performance compared to no augmentation.
    }
    \label{tab:vtt}
    \centering
    \begin{tabular}{|l|l|c|c|c|c|c|}
\hlineB{3}
\textbf{Dataset}                            & \textbf{Aug}          & \textbf{FPR} \scriptsize{[\%]$\uparrow$}       & \textbf{FNR} \scriptsize{[\%]$\uparrow$}       & \textbf{\textit{t} Test}    & \textbf{Diff}\scriptsize{$\downarrow$}                      & \textbf{KS Test}\\
\hlineB{3}
\multirow{4}{*}{ChestXray-14\phantom{0}}    & None                  & \textbf{\underline{48}}\phantom{\textdagger}   & \textbf{\underline{58}}\phantom{\textdagger}   & \cellcolor{gray!25}p=.497   & \phantom{-}0.12\phantom{\textdagger}                        & \cellcolor{gray!25}p=.869 \\
\cline{2-7}
                                            & ADA                   & 32\textdagger                                  & 47\textdagger                                  & \cellcolor{gray!25}p=.340   & \phantom{-}0.28\textdagger                                  & \cellcolor{gray!25}p=.549 \\
\cline{2-7}
                                            & APA                   &  34\textdagger                                 & 56\textdagger                                  & p=.082                      & \phantom{-}0.24\textdagger                                  & \cellcolor{gray!25}p=.717 \\
\cline{2-7}
                                            & DiffAug\phantom{0}    & \textbf{\underline{48}}\phantom{\textdagger}   & \textbf{\underline{58}}\phantom{\textdagger}   & \cellcolor{gray!25}p=.616   & \textbf{\underline{-0.16}}\phantom{\textdagger}             & \cellcolor{gray!25}p=.967 \\
\hlineB{3}
\multirow{4}{*}{SLIVER07}                   & None                  & 20\phantom{\textdagger}                        & \textbf{\underline{34}}\phantom{\textdagger}   & \cellcolor{gray!25}p=.424   & \phantom{-}0.68\phantom{\textdagger}                        & p$<$.001\\
\cline{2-7}
                                            & ADA                   & 24\phantom{\textdagger}                        & 30\textdagger                                  & \cellcolor{gray!25}p=.748   & \phantom{-}0.52\phantom{\textdagger}                        & p=.001\\
\cline{2-7}
                                            & APA                   & 10\textdagger                                  & 28\textdagger                                  & \cellcolor{gray!25}p=.232   & \phantom{-}0.82\textdagger                                  & p$<$.001\\
\cline{2-7}
                                            & DiffAug               & \textbf{\underline{34}}\phantom{\textdagger}   & 30\textdagger                                  & \cellcolor{gray!25}p=.825   & \phantom{-}\textbf{\underline{0.22}}\phantom{\textdagger}   & \cellcolor{gray!25}p=.717\\
\hlineB{3}
\multirow{4}{*}{MSD}                        & None                  & 58\phantom{\textdagger}                        & 48\phantom{\textdagger}                        & \cellcolor{gray!25}p=.543   & \phantom{-}0.08\phantom{\textdagger}                        & \cellcolor{gray!25}p$>$.999\\
\cline{2-7}
                                            & ADA                   & \textbf{\underline{66}}\phantom{\textdagger}   & 48\phantom{\textdagger}                        & \cellcolor{gray!25}p=.217   & -0.04\phantom{\textdagger}                                  & \cellcolor{gray!25}p$>$.999 \\
\cline{2-7}
                                            & APA                   & 46\textdagger                                  & 38\textdagger                                  & \cellcolor{gray!25}p=.587   & \phantom{-}0.04\phantom{\textdagger}                        & \cellcolor{gray!25}p$>$.999\\
\cline{2-7}
                                            & DiffAug               & 50\textdagger                                  & \textbf{\underline{54}}\phantom{\textdagger}   & \cellcolor{gray!25}p=.812   & \textbf{\underline{-0.08}}\phantom{\textdagger}             & \cellcolor{gray!25}p$>$.999 \\
\hlineB{3}
\multirow{4}{*}{ACDC}                       & None                  & 34\phantom{\textdagger}                        & 22\phantom{\textdagger}                        & \cellcolor{gray!25}p=.470   & \phantom{-}0.52\phantom{\textdagger}                        & p=.022\\
\cline{2-7}
                                            & ADA                   & 38\phantom{\textdagger}                        & \textbf{\underline{30}}\phantom{\textdagger}   & \cellcolor{gray!25}p=.653   & \phantom{-}0.38\phantom{\textdagger}                        & \cellcolor{gray!25}p=.112\\
\cline{2-7}
                                            & APA                   & 28\textdagger                                  & 22\phantom{\textdagger}                        & \cellcolor{gray!25}p=.707   & \phantom{-}0.46\phantom{\textdagger}                        & p=.003\\
\cline{2-7}
                                            & DiffAug               & \textbf{\underline{44}}\phantom{\textdagger}   & 16\textdagger                                  & p=.015                      & \phantom{-}\textbf{\underline{0.28}}\phantom{\textdagger}   & \cellcolor{gray!25}p=.112\\
\hlineB{3}
    \end{tabular}
\end{table}

\noindent consistent rankings among all ImageNet-based FDs, aligning closely with human judgment. 
In contrast, RadImageNet-based FDs exhibited volatility and diverged from human assessment. 
DiffAugment was the best-performing form of augmentation, generating hyper-realistic images on two datasets.

\textbf{ImageNet extractors aligned with human judgment.}
ImageNet-based FDs were consistent with one another in ranking generative models, except for on the MSD dataset, where human rankings were also inconsistent (see Figure \ref{fig:rank}). 
This consistency was reinforced by strong correlations between the FDs derived from InceptionV3 and all other ImageNet-based FDs (p$<$.001). 
Furthermore, the ImageNet-based FDs aligned with expert judgment (see Figure 1). 
On the ChestX-ray14 dataset, ImageNet-based FDs ranked generative models in the same order as the radiologists: DiffAugment, ADA, no augmentation, and APA. 
Particularly promising was the SwAV-based FD, which significantly correlated with human perception across all models (Pearson coefficient of .475 with the difference in average Likert ratings, p=.064).

\begin{table}
    \caption{ImageNet-based rFDs. 
    Column 1 lists each tested dataset, while Column 2 specifies the augmentation technique (Aug) utilized during model training: no augmentation (None), ADA, APA, and DiffAugment (DiffAug). 
    Columns 3-9 display the rFDs computed using seven ImageNet-trained feature extractors: InceptionV3 (Incept), ResNet50 (Res), InceptionResNetV2 (IRV2), DenseNet121 (Dense), SwAV, DINO, and Swin Transformer (Swin). 
    $\downarrow$ indicates that a lower value is preferable. 
    The underlined boldface type represents the best performance per dataset. 
    \textdagger\phantom{ }denotes decreased performance compared to no augmentation.
    }
    \label{tab:rfd_in}
    \centering
             \begin{tabular}{|l|l|c|c|c|c|c|c|c|}
             \hline
                           & & \multicolumn{7}{c|}{\textbf{Relative Fr\'{e}chet Distances} (ImageNet) \scriptsize{$\downarrow$}} \\
\cline{3-9}
\textbf{Dataset}  & \textbf{Aug} & Incept & Res & IRV2 & Dense & SwAV & DINO & Swin\\
\hlineB{3}
\multirow{4}{*}{ChestXray-14} & None    & 12.53\phantom{\textdagger}                                & 279.00\phantom{\textdagger}                                & \phantom{0}701.00\phantom{\textdagger}                      & \phantom{0}20.80\phantom{\textdagger}                      & \phantom{0}53.50\phantom{\textdagger}                      & \phantom{0}60.43\phantom{\textdagger}                      & \phantom{0}34.00\phantom{\textdagger}\\
\cline{2-9}
                              & ADA     & \phantom{0}8.90\phantom{\textdagger}                      & 237.00\phantom{\textdagger}                                & \phantom{0}576.00\phantom{\textdagger}                      & \phantom{0}15.55\phantom{\textdagger}                      & \phantom{0}33.00\phantom{\textdagger}                      & \phantom{0}37.81\phantom{\textdagger}                      & \phantom{0}26.36\phantom{\textdagger}\\
\cline{2-9}
                              & APA     & 17.58\textdagger                                          & 334.00\textdagger                                          & 1004.50\textdagger                                          & \phantom{0}39.85\textdagger                                & \phantom{0}66.00\textdagger                                & \phantom{0}82.23\textdagger                                & \phantom{0}54.21\textdagger\\
\cline{2-9}
                              & DiffAug & \phantom{0}\textbf{\underline{7.68}}\phantom{\textdagger} & \textbf{\underline{146.00}}\phantom{\textdagger}           & \phantom{0}\textbf{\underline{441.00}}\phantom{\textdagger} & \phantom{0}\textbf{\underline{13.25}}\phantom{\textdagger} & \phantom{0}\textbf{\underline{25.00}}\phantom{\textdagger} & \phantom{0}\textbf{\underline{34.51}}\phantom{\textdagger} & \phantom{0}\textbf{\underline{22.79}}\phantom{\textdagger}\\
\hlineB{3}
\multirow{4}{*}{SLIVER07}     & None    & \phantom{0}1.48\phantom{\textdagger}                      & \phantom{00}7.90\phantom{\textdagger}                      & \phantom{00}12.98\phantom{\textdagger}                      & \phantom{00}2.59\phantom{\textdagger}                      & \phantom{00}8.28\phantom{\textdagger}                      & \phantom{00}6.12\phantom{\textdagger}                      & \phantom{00}6.07\phantom{\textdagger}\\
\cline{2-9}
                              & ADA     & \phantom{0}1.24\phantom{\textdagger}                      & \phantom{00}7.35\phantom{\textdagger}                      & \phantom{00}11.71\phantom{\textdagger}                      & \phantom{00}1.95\phantom{\textdagger}                      & \phantom{00}6.86\phantom{\textdagger}                      & \phantom{00}4.57\phantom{\textdagger}                      & \phantom{00}6.22\textdagger\\
\cline{2-9}
                              & APA     & \phantom{0}1.37\phantom{\textdagger}                      & \phantom{00}7.33\phantom{\textdagger}                      & \phantom{00}11.96\phantom{\textdagger}                      & \phantom{00}2.36\phantom{\textdagger}                      & \phantom{00}7.79\phantom{\textdagger}                      & \phantom{00}5.59\phantom{\textdagger}                      & \phantom{00}5.43\phantom{\textdagger}\\
\cline{2-9}
                              & DiffAug & \phantom{0}\textbf{\underline{0.78}}\phantom{\textdagger} & \phantom{00}\textbf{\underline{3.25}}\phantom{\textdagger} & \phantom{000}\textbf{\underline{5.99}}\phantom{\textdagger} & \phantom{00}\textbf{\underline{1.24}}\phantom{\textdagger} & \phantom{00}\textbf{\underline{5.26}}\phantom{\textdagger} & \phantom{00}\textbf{\underline{3.07}}\phantom{\textdagger} & \phantom{00}\textbf{\underline{4.77}}\phantom{\textdagger}\\
\hlineB{3}
\multirow{4}{*}{MSD}          & None    & 37.32\phantom{\textdagger}                                & \phantom{0}63.13\phantom{\textdagger}                      & \phantom{00}61.18\phantom{\textdagger}                      & 170.38\phantom{\textdagger}                                & 142.50\phantom{\textdagger}                                & \textbf{\underline{108.39}}\phantom{\textdagger}           & 504.47\phantom{\textdagger}\\
\cline{2-9}
                              & ADA     & \textbf{\underline{36.84}}\phantom{\textdagger}           & \phantom{0}\textbf{\underline{62.50}}\phantom{\textdagger} & \phantom{00}\textbf{\underline{58.88}}\phantom{\textdagger} & \textbf{\underline{141.63}}\phantom{\textdagger}           & 305.00\textdagger                                          & 121.90\textdagger                                          & 308.59\phantom{\textdagger}\\
\cline{2-9}
                              & APA     & 43.63\textdagger                                          & \phantom{0}70.00\textdagger                                & \phantom{00}81.76\textdagger                                & 145.13\phantom{\textdagger}                                & \textbf{\underline{122.50}}\phantom{\textdagger}           & 126.47\textdagger                                          & 196.65\phantom{\textdagger}\\
\cline{2-9}
                              & DiffAug & 46.32\textdagger                                          & 125.50\textdagger                                          & \phantom{00}79.88\textdagger                                & 170.38\phantom{\textdagger}                                & 825.00\textdagger                                          & 138.11\textdagger                                          & \textbf{\underline{175.12}}\phantom{\textdagger}\\
\hlineB{3}
\multirow{4}{*}{ACDC}         & None    & 49.67\phantom{\textdagger}                                & \phantom{0}86.48\phantom{\textdagger}                      & \phantom{0}121.14                                           & \phantom{0}87.46\phantom{\textdagger}                      & 118.00\phantom{\textdagger}                                & 140.15\phantom{\textdagger}                                & 111.07\phantom{\textdagger}\\
\cline{2-9}
                              & ADA     & 20.99\phantom{\textdagger}                                & \phantom{0}31.66\phantom{\textdagger}                      & \phantom{00}49.94                                           & \phantom{0}35.95\phantom{\textdagger}                      & \phantom{0}76.40\phantom{\textdagger}                      & \phantom{0}65.52\phantom{\textdagger}                      & \phantom{0}61.49\phantom{\textdagger}\\
\cline{2-9}
                              & APA     & 31.15\phantom{\textdagger}                                & \phantom{0}54.35\phantom{\textdagger}                      & \phantom{00}76.47                                           & \phantom{0}56.68\phantom{\textdagger}                      & \phantom{0}90.60\phantom{\textdagger}                      & \phantom{0}87.69\phantom{\textdagger}                      & \phantom{0}72.10\phantom{\textdagger}\\
\cline{2-9}
                              & DiffAug & \textbf{\underline{15.87}}\phantom{\textdagger}           & \phantom{0}\textbf{\underline{23.58}}\phantom{\textdagger} & \phantom{00}\textbf{\underline{40.60}}                      & \phantom{0}\textbf{\underline{27.20}}\phantom{\textdagger} & \phantom{0}\textbf{\underline{71.00}}\phantom{\textdagger} & \phantom{0}\textbf{\underline{50.47}}\phantom{\textdagger} & \phantom{0}\textbf{\underline{47.23}}\phantom{\textdagger}\\
\hlineB{3}
    \end{tabular}
\end{table}

\textbf{RadImageNet extractors were volatile.}
RadImageNet-based FDs produced inconsistent rankings that were contrary to expert judgment. 
Notably, on the SLIVER07 dataset, RadImageNet-based FDs ranked DiffAugment as one of the poorest-performing models. 
However, all measures of human judgment identified DiffAugment as the best-performing model (see Figure \ref{fig:rank}).
This discrepancy is especially concerning considering RadImageNet's inclusion of approximately 300,000 CT scans.
On the ChestX-ray14 dataset, the FD derived from a RadImageNet-trained InceptionV3 network ranked the model without augmentation as the best performing.
In contrast, a thoracic radiologist observed that both the APA and no augmentation models generated multiple radiographs with obviously distorted anatomy.
Conversely, the weaknesses of the DiffAugment and ADA models were more subtle, with mistakes in support devices and central lines.

\begin{table}
    \caption{RadImageNet-based rFDs. 
    Column 1 lists each tested dataset, while Column 2 specifies the augmentation technique (Aug) utilized during model training: no augmentation (None), ADA, APA, and DiffAugment (DiffAug). 
    Columns 3-6 display the rFDs computed using four RadImageNet-trained feature extractors: InceptionV3, ResNet50, InceptionResNetV2 (IRV2), and DenseNet121. 
    $\downarrow$ indicates that a lower value is preferable. 
    The underlined boldface type represents the best performance per dataset. 
    \textdagger\phantom{ }denotes decreased performance compared to no augmentation.
    }
    \label{tab:rfd_rin}
    \centering
             \begin{tabular}{|l|l|c|c|c|c|}
             \hline
                           & & \multicolumn{4}{c|}{\phantom{ }\textbf{Relative Fr\'{e}chet Distances} (RadImageNet) \scriptsize{$\downarrow$}\phantom{ }} \\
\cline{3-6}
\textbf{Dataset}                         & \textbf{Aug} & \phantom{ }InceptionV3\phantom{ } & \phantom{ }ResNet50\phantom{ } & IRV2 & DenseNet121\\
\hlineB{3}
\multirow{4}{*}{ChestXray-14}\phantom{ } & None               & \phantom{0}\textbf{\underline{140.00}}\phantom{\textdagger} & \phantom{00}75.00\phantom{\textdagger}                       & \phantom{0}\textbf{\underline{80.00}}\phantom{\textdagger} & \phantom{0}40.00\phantom{\textdagger}\\
\cline{2-6}
                                         & ADA                & \phantom{0}660.00\textdagger                                & \phantom{0}135.00\textdagger                                & 190.00\textdagger                                          & \phantom{0}80.00\textdagger\\
\cline{2-6}
                                         & APA                & \phantom{0}280.00\textdagger                                & \phantom{00}65.00\phantom{\textdagger}                      & \phantom{0}\textbf{\underline{80.00}}\phantom{\textdagger} & \phantom{0}80.00\textdagger\\
\cline{2-6}
                                         & DiffAug\phantom{0} & \phantom{0}280.00\textdagger                                & \phantom{00}\textbf{\underline{50.00}}\phantom{\textdagger} & \phantom{0}90.00\textdagger                                & \phantom{0}\textbf{\underline{30.00}}\phantom{\textdagger}\\
\hlineB{3}
\multirow{4}{*}{SLIVER07}                & None               & \phantom{000}3.67\phantom{\textdagger}                      & \phantom{000}3.14\phantom{\textdagger}                      & \phantom{00}6.00\phantom{\textdagger}                      & \phantom{00}4.33\phantom{\textdagger}\\
\cline{2-6}
                                         & ADA                & \phantom{000}\textbf{\underline{1.89}}\phantom{\textdagger} & \phantom{000}\textbf{\underline{1.86}}\phantom{\textdagger} & \phantom{00}3.75\phantom{\textdagger}                      & \phantom{00}\textbf{\underline{2.33}}\phantom{\textdagger}\\
\cline{2-6}
                                         & APA                & \phantom{000}2.22\phantom{\textdagger}                      & \phantom{000}\textbf{\underline{1.86}}\phantom{\textdagger} & \phantom{00}\textbf{\underline{3.00}}\phantom{\textdagger} & \phantom{00}2.67\phantom{\textdagger}\\
\cline{2-6}
                                         & DiffAug            & \phantom{000}4.67\textdagger                                & \phantom{000}3.29\textdagger                                & \phantom{00}5.50\phantom{\textdagger}                      & \phantom{00}4.67\textdagger\\
\hlineB{3}
\multirow{4}{*}{MSD}                     & None               & \phantom{00}53.00\phantom{\textdagger}                      & \phantom{00}32.50\phantom{\textdagger}                      & \phantom{0}\textbf{\underline{32.50}}\phantom{\textdagger} & \phantom{0}\textbf{\underline{40.00}}\phantom{\textdagger}\\
\cline{2-6}
                                         & ADA                & \phantom{00}\textbf{\underline{36.00}}\phantom{\textdagger} & \phantom{00}\textbf{\underline{27.5}}\phantom{\textdagger}  & \phantom{0}37.50\textdagger                                & \phantom{0}60.00\textdagger\\
\cline{2-6}
                                         & APA                & \phantom{00}54.00\textdagger                                & \phantom{00}32.50\phantom{\textdagger}                      & \phantom{0}40.00\textdagger                                & \phantom{0}\textbf{\underline{40.00}}\phantom{\textdagger}\\
\cline{2-6}
                                         & DiffAug            & 1551.00\textdagger                                          & 1105.00\textdagger                                          & 350.00\textdagger                                          & 615.00\textdagger\\
\hlineB{3}
\multirow{4}{*}{ACDC}                    & None               & \phantom{00}26.64\phantom{\textdagger}                      & \phantom{00}19.00\phantom{\textdagger}                                  & \phantom{0}20.33\phantom{\textdagger}                      & \phantom{0}32.50\phantom{\textdagger}\\
\cline{2-6}
                                         & ADA                & \phantom{00}\textbf{\underline{10.18}}\phantom{\textdagger} & \phantom{000}9.25\phantom{\textdagger}                        & \phantom{00}\textbf{\underline{9.67}}\phantom{\textdagger} & \phantom{0}13.00\phantom{\textdagger}\\
\cline{2-6}
                                         & APA                & \phantom{00}14.09\phantom{\textdagger}                      & \phantom{000}\textbf{\underline{8.75}}\phantom{\textdagger}   & \phantom{0}11.67\phantom{\textdagger}                      & \phantom{0}17.50\phantom{\textdagger}\\
\cline{2-6}
                                         & DiffAug            & \phantom{00}12.09\phantom{\textdagger}                      & \phantom{00}15.25\phantom{\textdagger}                                  & \phantom{00}\textbf{\underline{9.67}}\phantom{\textdagger} & \phantom{0}\textbf{\underline{10.50}}\phantom{\textdagger}\\
\hlineB{3}
    \end{tabular}
\end{table}

\textbf{APA and ADA demonstrated varied performance.} 
Although APA was designed to enhance image quality in limited data domains such as medical imaging, it unexpectedly reduced the perceptual quality of the generated images (p=.012), leading to an 18\% reduction in the FPR on average. 
While ADA outperformed APA (p=.050), it did not significantly affect participants' ability to differentiate real from generated images (p$>$.999). 
Despite both techniques underperforming in the VTTs, they improved the rFDs for the SLIVER07 (p=.025 ADA, p=.016 APA) and ACDC (p=.003 ADA, p=.004 APA) datasets.

\textbf{DiffAugment created hyper-realistic images}.
DiffAugment outperformed the other augmentation techniques across all FDs (p=.092 ADA, p=.059 APA).
This result held for each dataset except MSD, where model training had diverged. 
DiffAugment was the only form of augmentation to significantly enhance perceptual quality (p=.001), resulting in an 81\% reduction in the average difference between mean Likert ratings.
Participants rated images from DiffAugment-based models as more realistic than those from both the ChestX-ray14 and MSD datasets. 
Additionally, Likert ratings for real and generated images from all DiffAugment-based models did not differ significantly (p=.793), suggesting that participants perceived them as equivalent.

\section{Conclusion}

Our study challenges prevailing assumptions by providing novel evidence that medical image-trained feature extractors do not inherently improve FDs for synthetic medical imaging evaluation;
instead, they may compromise metric consistency and alignment with human judgment, even on in-domain data. 
The emerging practice of employing privately trained, medical image-based feature extractors to benchmark new generative algorithms is concerning, as it allows algorithm designers to shape evaluation metrics, potentially introducing biases. 
Additionally, the efficacy of these FDs often remains inadequately evaluated and unverifiable. 
We advocate for the comprehensive evaluation and public release of all FDs used in benchmarking generative medical imaging models.

\subsubsection{Acknowledgments}

Research reported in this publication was supported in part by resources of the Image Guided Cancer Therapy Research Program at The University of Texas MD Anderson Cancer Center, by a generous gift from the Apache Corporation, by the National Institutes of Health/NCI under award number P30CA016672, and by the Tumor Measurement Initiative through the MD Anderson Strategic Initiative Development Program (STRIDE). 
We thank the NIH Clinical Center for the ChestX-ray14 dataset, the StudioGAN authors \cite{Kang2023} for their FD implementations, Vikram Haheshri and Oleg Igoshin for the discussion that led to the hypothesis testing contribution, Erica Goodoff - Senior Scientific Editor in the Research Medical Library at The University of Texas MD Anderson Cancer Center - for editing this article, and Xinyue Zhang and Caleb O'Connor for their comments when reviewing the manuscript.
GPT4 was used in the proofreading stage of this manuscript.

\bibliographystyle{splncs04}

\newpage

\noindent{\large \textbf{Appendix}}
\begin{table}[h]
    \caption{ImageNet-based FDs. 
    Column 1 lists each tested dataset, while Column 2 specifies the augmentation technique (Aug) utilized during model training: no augmentation (None), ADA, APA, and DiffAugment (DiffAug). 
    Columns 3-9 display the FDs computed using seven ImageNet-trained feature extractors: InceptionV3 (Incept), ResNet50 (Res), InceptionResNetV2 (IRV2), DenseNet121 (Dense), SwAV, DINO, and Swin Transformer (Swin). 
    $\downarrow$ indicates that a lower value is preferable. 
    The underlined boldface type represents the best performance per dataset. 
    \textdagger\phantom{ }denotes decreased performance compared to no augmentation.
    }
    \label{tab:fd_in}
    \centering
\begin{tabular}{|l|l|c|c|c|c|c|c|c|}
             \hline
                           & & \multicolumn{7}{c|}{\textbf{Fr\'{e}chet Distances} (ImageNet) \scriptsize{$\downarrow$}} \\
\cline{3-9}
\textbf{Dataset}  & \textbf{Aug} & Incept & Res & IRV2 & Dense & SwAV & DINO & Swin\\
\hlineB{3}
\multirow{4}{*}{ChestXray-14}\phantom{0}   & None                 & \phantom{0}5.01\phantom{\textdagger}            & \phantom{0}4.16\phantom{\textdagger}            &\phantom{0}2.79\phantom{\textdagger}&14.02\phantom{\textdagger}& 1.07\phantom{\textdagger}                & \phantom{0}299.13\phantom{\textdagger}            & \phantom{00}4.76\phantom{\textdagger}\\
\cline{2-9}
                                           & ADA                  & \phantom{0}3.56\phantom{\textdagger}            & \phantom{0}3.11\phantom{\textdagger}            &\phantom{0}2.37\phantom{\textdagger}&11.52\phantom{\textdagger}& 0.66\phantom{\textdagger}                & \phantom{0}187.16\phantom{\textdagger}            & \phantom{00}3.69\phantom{\textdagger}\\
\cline{2-9}
                                           & APA                  & \phantom{0}7.03\textdagger                      & \phantom{0}7.97\textdagger                      &\phantom{0}3.34\textdagger&20.09\textdagger& 1.32\textdagger                          & \phantom{0}407.05\textdagger                      & \phantom{00}7.49\textdagger                      \\
\cline{2-9}
                                           & DiffAug\phantom{0}   & \phantom{0}\textbf{\underline{3.07}}\phantom{\textdagger}   & \phantom{0}\textbf{\underline{2.65}}\phantom{\textdagger}   &\phantom{0}\textbf{\underline{1.46}}\phantom{\textdagger}&\phantom{0}\textbf{\underline{8.82}}\phantom{\textdagger}& \textbf{\underline{0.50}}\phantom{\textdagger}       & \phantom{0}\textbf{\underline{170.84}}\phantom{\textdagger}   & \phantom{00}\textbf{\underline{3.19}}\phantom{\textdagger}\\
\hlineB{3}
\multirow{4}{*}{SLIVER07}                  & None                 & \phantom{0}8.72\phantom{\textdagger}            & \phantom{0}9.04\phantom{\textdagger}            &\phantom{0}4.74\phantom{\textdagger}&14.02\phantom{\textdagger}& 2.40\phantom{\textdagger}                & \phantom{0}640.32\phantom{\textdagger}            & \phantom{0}30.37\phantom{\textdagger}            \\
\cline{2-9}
                                           & ADA                  & \phantom{0}7.34\phantom{\textdagger}            & \phantom{0}6.79\phantom{\textdagger}            &\phantom{0}4.41\phantom{\textdagger}&12.65\phantom{\textdagger}& 1.99\phantom{\textdagger}                & \phantom{0}478.61\phantom{\textdagger}            & \phantom{0}31.09\textdagger                   \\
\cline{2-9}
                                           & APA                  & \phantom{0}8.07\phantom{\textdagger}            & \phantom{0}8.22\phantom{\textdagger}            &\phantom{0}4.40\phantom{\textdagger}&12.92\phantom{\textdagger}& 2.26\phantom{\textdagger}                & \phantom{0}585.60\phantom{\textdagger}            & \phantom{0}27.17\phantom{\textdagger}\\
\cline{2-9}
                                           & DiffAug              & \phantom{0}\textbf{\underline{4.62}}\phantom{\textdagger}   & \phantom{0}\textbf{\underline{4.33}}\phantom{\textdagger}   &\phantom{0}\textbf{\underline{1.95}}\phantom{\textdagger}&\phantom{0}\textbf{\underline{6.47}}\phantom{\textdagger}& \textbf{\underline{1.53}}\phantom{\textdagger}       & \phantom{0}\textbf{\underline{321.32}}\phantom{\textdagger}   & \phantom{0}\textbf{\underline{23.83}}\phantom{\textdagger}\\
\hlineB{3}
\multirow{4}{*}{MSD}                       & None                 & \phantom{0}7.09\phantom{\textdagger}            & \phantom{0}5.05\phantom{\textdagger}            &10.40\phantom{\textdagger}&\phantom{0}\textbf{\underline{9.43}}\phantom{\textdagger}& 0.57\phantom{\textdagger}                & \phantom{0}\textbf{\underline{422.74}}\phantom{\textdagger}   & \phantom{0}85.76\phantom{\textdagger}\\
\cline{2-9}
                                           & ADA                  & \phantom{0}\textbf{\underline{7.00}}\phantom{\textdagger}   & \phantom{0}\textbf{\underline{5.00}}\phantom{\textdagger}   &\textbf{\underline{10.01}}\phantom{\textdagger}&11.33\textdagger& 1.22\textdagger                          & \phantom{0}475.41\textdagger                      & \phantom{0}52.46\phantom{\textdagger}\\
\cline{2-9}
                                           & APA                  & \phantom{0}8.29\textdagger                      & \phantom{0}5.60\textdagger                      &13.90\textdagger&11.61\textdagger& \textbf{\underline{0.49}}\phantom{\textdagger}       & \phantom{0}493.23\textdagger                      & \phantom{0}33.43\phantom{\textdagger}\\
\cline{2-9}
                                           & DiffAug              & \phantom{0}8.80\textdagger                      & 10.04\textdagger                                &13.58\textdagger&13.63\textdagger& 3.30\textdagger                          & \phantom{0}538.61\textdagger                      & \phantom{0}\textbf{\underline{29.77}}\phantom{\textdagger}\\
\hlineB{3}
\multirow{4}{*}{ACDC}                      & None                 & 67.05\phantom{\textdagger}                      & 51.60\phantom{\textdagger}                      &73.51&87.22\phantom{\textdagger}& 5.90\phantom{\textdagger}                & 2888.53\phantom{\textdagger}                      & 127.74\phantom{\textdagger}\\
\cline{2-9}
                                           & ADA                  & 28.34\phantom{\textdagger}                      & 21.21\phantom{\textdagger}                      &26.91&35.96\phantom{\textdagger}& 3.82\phantom{\textdagger}                & 1350.34\phantom{\textdagger}                      & \phantom{0}70.71\phantom{\textdagger}\\
\cline{2-9}
                                           & APA                  & 42.05\phantom{\textdagger}                      & 33.44\phantom{\textdagger}                      &46.20&55.06\phantom{\textdagger}& 4.53\phantom{\textdagger}                & 1807.38\phantom{\textdagger}                      & \phantom{0}82.91\phantom{\textdagger}\\
\cline{2-9}
                                           & DiffAug              & \textbf{\underline{21.42}}\phantom{\textdagger}             & \textbf{\underline{16.05}}\phantom{\textdagger}             &\textbf{\underline{20.04}}&\textbf{\underline{29.23}}\phantom{\textdagger}& \textbf{\underline{3.55}}\phantom{\textdagger}       & \textbf{\underline{1040.24}}\phantom{\textdagger}             & \phantom{0}\textbf{\underline{54.31}}\phantom{\textdagger}\\
\hlineB{3}
    \end{tabular}
\end{table}

\begin{table}
    \caption{RadImageNet-based FDs. 
    Column 1 lists each tested dataset, while Column 2 specifies the augmentation technique (Aug) utilized during model training: no augmentation (None), ADA, APA, and DiffAugment (DiffAug). 
    Columns 3-6 display the FDs computed using four RadImageNet-trained feature extractors: InceptionV3, ResNet50, InceptionResNetV2 (IRV2), and DenseNet121. 
    $\downarrow$ indicates that a lower value is preferable. 
    The underlined boldface type represents the best performance per dataset. 
    \textdagger\phantom{ }denotes decreased performance compared to no augmentation.
    }
    \label{tab:fd_rin}
    \centering
\begin{tabular}{|l|l|c|c|c|c|}
             \hline
                           & & \multicolumn{4}{c|}{\phantom{ }\textbf{Fr\'{e}chet Distances} (RadImageNet) \scriptsize{$\downarrow$}\phantom{ }} \\
\cline{3-6}
\textbf{Dataset}                         & \textbf{Aug}           & \phantom{ }InceptionV3\phantom{ }              & \phantom{ }ResNet50\phantom{ }                 & IRV2                                           & DenseNet121\\
\hlineB{3}
\multirow{4}{*}{ChestXray-14}\phantom{0}   & None                 & \textbf{\underline{0.03}}\phantom{\textdagger} & 0.15\phantom{\textdagger}                      & \textbf{\underline{0.08}}\phantom{\textdagger} & 0.04\phantom{\textdagger} \\
\cline{2-6}
                                           & ADA                  & 0.13\textdagger                                & 0.27\textdagger                                & 0.19\textdagger                                & 0.08\textdagger \\
\cline{2-6}
                                           & APA                  & 0.06\textdagger                                & 0.13\phantom{\textdagger}                      & \textbf{\underline{0.08}}\phantom{\textdagger} & 0.08\textdagger \\
\cline{2-6}
                                           & DiffAug\phantom{0}   & 0.06\textdagger                                & \textbf{\underline{0.10}}\phantom{\textdagger} & 0.09\textdagger                                & \textbf{\underline{0.03}}\phantom{\textdagger} \\
\hlineB{3}
\multirow{4}{*}{SLIVER07}                  & None                 & 0.07\phantom{\textdagger}                      & 0.22\phantom{\textdagger}                      & 0.24\phantom{\textdagger}                      & 0.13\phantom{\textdagger} \\
\cline{2-6}
                                           & ADA                  & \textbf{\underline{0.03}}\phantom{\textdagger} & \textbf{\underline{0.13}}\phantom{\textdagger} & 0.15\phantom{\textdagger}                      & \textbf{\underline{0.07}}\phantom{\textdagger} \\
\cline{2-6}
                                           & APA                  & 0.04\phantom{\textdagger}                      & \textbf{\underline{0.13}}\phantom{\textdagger} & \textbf{\underline{0.12}}\phantom{\textdagger} & 0.08\phantom{\textdagger} \\
\cline{2-6}
                                           & DiffAug              & 0.08\textdagger                                & 0.23\textdagger                                & 0.22\phantom{\textdagger}                      & 0.14\textdagger \\
\hlineB{3}
\multirow{4}{*}{MSD}                       & None                 & 0.05\phantom{\textdagger}                      & 0.13\phantom{\textdagger}                      & \textbf{\underline{0.13}}\phantom{\textdagger} & \textbf{\underline{0.08}}\phantom{\textdagger} \\
\cline{2-6}
                                           & ADA                  & \textbf{\underline{0.04}}\phantom{\textdagger} & \textbf{\underline{0.11}}\phantom{\textdagger} & 0.15\textdagger                                & 0.13\textdagger \\
\cline{2-6}
                                           & APA                  & 0.05\phantom{\textdagger}                      & 0.13\phantom{\textdagger}                      & 0.16\textdagger                                & \textbf{\underline{0.08}}\phantom{\textdagger} \\
\cline{2-6}
                                           & DiffAug              & 1.55\textdagger                                & 4.42\textdagger                                & 1.40\textdagger                                & 1.23\textdagger \\
\hlineB{3}
\multirow{4}{*}{ACDC}                      & None                 & 0.29\phantom{\textdagger}                      & 0.76\phantom{\textdagger}                      & 0.61\phantom{\textdagger}                      & 0.65\phantom{\textdagger} \\
\cline{2-6}
                                           & ADA                  & \textbf{\underline{0.11}}\phantom{\textdagger} & 0.37\phantom{\textdagger}                      & \textbf{\underline{0.29}}\phantom{\textdagger} & 0.26\phantom{\textdagger} \\
\cline{2-6}
                                           & APA                  & 0.15\phantom{\textdagger}                      & \textbf{\underline{0.35}}\phantom{\textdagger} & 0.35\phantom{\textdagger}                      & 0.35\phantom{\textdagger} \\
\cline{2-6}
                                           & DiffAug              & 0.13\phantom{\textdagger}                      & 0.61\phantom{\textdagger}                      & \textbf{\underline{0.29}}\phantom{\textdagger} & \textbf{\underline{0.21}}\phantom{\textdagger} \\
\hlineB{3}
    \end{tabular}
\end{table}

\end{document}